\documentclass{article} % For LaTeX2e
\usepackage{iclr2027_conference,times}

\usepackage{amsmath,amsfonts,bm}

\def\eqref#1{equation~\ref{#1}}
\def\1{\bm{1}}

\DeclareMathAlphabet{\mathsfit}{\encodingdefault}{\sfdefault}{m}{sl}
\SetMathAlphabet{\mathsfit}{bold}{\encodingdefault}{\sfdefault}{bx}{n}

\usepackage{hyperref}
\usepackage{url}

\usepackage{graphicx}
\usepackage{subcaption}
\usepackage{booktabs}
\usepackage{makecell}
\usepackage{algorithm}
\usepackage{algpseudocode}   
\usepackage{caption}
\usepackage[table]{xcolor}
\usepackage{xspace}
\usepackage{wrapfig}
\newcommand{\method}{\text{ReactiveGWM}\xspace}

\title{ReactiveGWM: Flexible Control and \\ NPC Reactivity in Game World Models}
\author{%
  \textbf{Zeqing Wang}$^{1,2,\clubsuit}$\quad
  \textbf{Danze Chen}$^{1,2,\clubsuit}$\quad
  \textbf{Zhaohu Xing}$^{4}$\quad
  \textbf{Zizhao Tong}$^{1,5,\clubsuit}$\\[3pt]
  \textbf{Yinhan Zhang}$^{1,6,\clubsuit}$\quad
  \textbf{Xingyi Yang}$^{3}$%
  \thanks{Corresponding authors.}\quad
  \textbf{Yeying Jin}$^{1,2,\diamondsuit}$%
  \footnotemark[1]\\[6pt]
  $^{1}$Tencent, $^{2}$National University of Singapore, $^{3}$The Hong Kong Polytechnic University,\\
  $^{4}$The Hong Kong University of Science and Technology (Guangzhou),\\
  $^{5}$University of Chinese Academy of Sciences,\\
  $^{6}$The Hong Kong University of Science and Technology\\[4pt]
  \texttt{zeqing.wang@u.nus.edu}\quad
  \texttt{xingyi.yang@polyu.edu.hk}\\
  \texttt{jinyeying@u.nus.edu}%
}
\iclrfinalcopy % Uncomment for camera-ready version, but NOT for submission.
\begin{document}

\maketitle
\ificlrfinal
  % Internship footnote
  \begingroup
    \renewcommand{\thefootnote}{\ensuremath{\clubsuit}}
    \footnotetext[0]{Work done during an internship at Tencent.}
  \endgroup

  % Project leader footnote
  \begingroup
    \renewcommand{\thefootnote}{\ensuremath{\diamondsuit}}
    \footnotetext[0]{Project leader.}
  \endgroup
\fi

\begin{figure}[H]
    \centering
    \includegraphics[width=\textwidth]{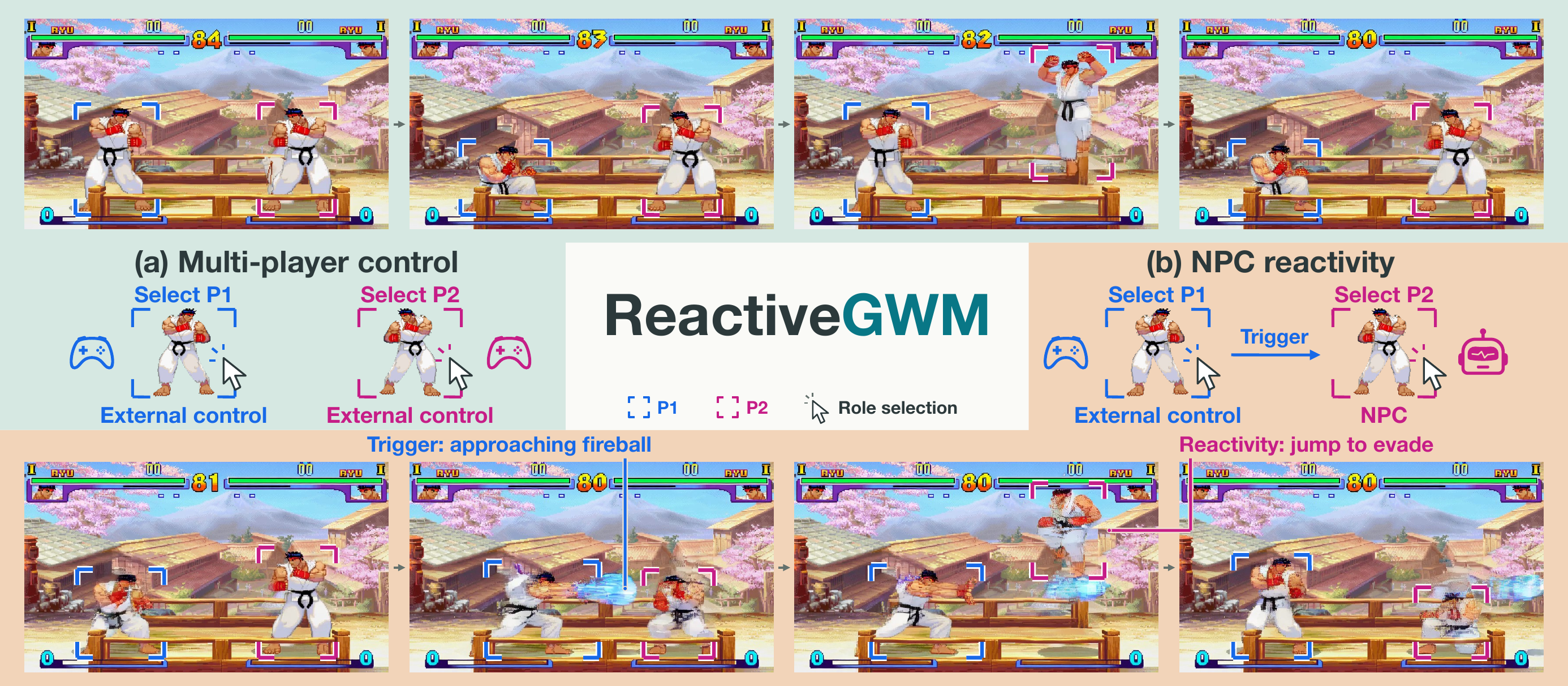}
    \caption{\method enables flexible role assignment and reactive game generation: (a) both characters follow external actions; (b) P1 is externally controlled, while P2 acts as an NPC and jumps to evade an approaching fireball.}
    \label{fig:teaser}
\end{figure}

\begin{abstract}

Existing game world models typically adopt role-specific interactions, where player and NPC roles are bound to fixed characters. This limits their flexibility in multi-character games, where different characters may receive external control while NPCs must react to interactions triggered by players. This setting raises two key challenges: \textit{how to flexibly assign control roles to individual characters, and how to support direct player control and reactive NPC behavior within a unified model}. These challenges are particularly pronounced in shared-view 2D games, where multiple, potentially visually identical characters share the same viewpoint, making camera cues insufficient to distinguish their roles.
To address these challenges, we introduce \textbf{\method}, a reactive game world model that flexibly assigns control modes at initialization and jointly simulates externally controlled players and reactive NPCs. Specifically, \method introduces \textit{Spatial Role Binding}, which grounds learned character handles to their corresponding regions in the initial frame using instance masks. Building on these handles, \textit{Unified Agency Conditioning} unifies heterogeneous control signals across characters by encoding player actions and conditional NPC rules into character-specific control-token groups. Each group is then bound to its corresponding character handle, enabling the model to apply each control signal to its designated character. 
Meanwhile, causal self-attention restricts temporal context to the current and preceding latent frames when generating player actions and NPC responses.
Experiments on two multi-character 2D games demonstrate that \method supports flexible character control across different player/NPC role assignments while jointly generating accurate player-controlled behaviors and reactive NPC responses, enabling more configurable and richer multi-character interactions. Anonymous project page is available at \url{https://inv-wzq.github.io/ReactiveGWM/}.

\end{abstract}
\section{Introduction}
Recent advancements in world models~\citep{world_model,genie3} have established a new paradigm for simulating complex and interactive environments. By learning environment dynamics from large-scale offline gameplay videos, this paradigm naturally extends to game world models~\citep{Lingbot-world,matrixgame3,WanToFight}, opening up new possibilities for interactive game generation.

To enable such interactive games, a key requirement is to model reactive interactions between players and Non-Player Characters (NPCs), where player inputs directly control selected characters, while player actions can automatically trigger corresponding NPC responses. In multi-character games, such interactions may involve different roles across characters. A character may be selected for external control in one rollout while serving as a reactive NPC in another. This flexible role assignment enables more diverse interaction patterns and greater gameplay flexibility.
%For example, two characters may swap their roles across rollouts, with one acting as the player and the other as a reactive NPC in one rollout, and vice versa in another. 
%Therefore, multi-character interaction requires the model to determine not only which character is controlled, but also which characters act as reactive NPCs and whom they should react to.

However, most existing game world models~\citep{WanToFight,Incantation,WorldMind} commonly adopt a fixed interaction structure: external actions are bound to predefined player characters, while the remaining characters retain fixed NPC roles. We refer to this setting as \textit{role-specific interaction}. It simplifies action grounding because the identity of the externally controlled character is known in advance. Hence, it restricts the range of interactions that a model can represent when different characters need to assume different control roles.

Moving beyond such fixed interaction structures introduces two closely related challenges. \textbf{First}, external actions must be grounded to the intended character rather than being tied to a predefined player/NPC role. \textbf{Second}, the model must support direct player control and reactive NPC behavior within a unified framework. These challenges are particularly pronounced in shared-view 2D games~\citep{sf3}, where multiple characters appear simultaneously under the same viewpoint, providing no camera-based cues for identifying their roles~\citep{Solaris}. Moreover, characters may have similar or even identical appearances, making visual identity alone insufficient to determine which character should receive external control and which should exhibit reactive behavior. Therefore, the model must explicitly associate external actions and reactive behaviors with individual characters while jointly modeling their dynamics within the same visual space.

To address these challenges, we introduce \textbf{\method}, a reactive game world model for multi-character 2D games. At rollout initialization, users assign each visible character either external control or an NPC behavior rule. The model then binds actions to the selected player characters and simulates the remaining NPCs according to their conditional rules. These role assignments remain fixed within a rollout, while the generated interaction evolves in response to player actions.

Specifically, \method comprises two complementary components. \textit{Spatial Role Binding} associates each character with a learned handle and injects mask-weighted handles into the initial-frame video tokens. This establishes a correspondence between a character's spatial location and the controls assigned to it. \textit{Unified Agency Conditioning} represents player actions and conditional NPC rules as control-token groups, binding each group to its character through the same handle. For NPCs, these rules encode trigger-response relationships that specify how they should react to particular player behaviors. Since NPC reactions are inherently conditioned on preceding player behaviors, we adopt causal self-attention to preserve this temporal dependency and prevent future video context from interfering with trigger interpretation.
In our experiments, we evaluate \method on two multi-character 2D games with varying player/NPC role assignments and trigger-response interactions. The results demonstrate that \method enables flexible character control while jointly modeling reactive NPC behavior, highlighting its potential to support richer and more configurable interactions between players and NPCs.

In summary, our main contributions are as follows:
\begin{itemize}
\item We formulate multi-character reactive interaction beyond role-specific game world models, allowing different characters to assume externally controlled or reactive NPC roles.

\item We introduce \textit{Spatial Role Binding} and \textit{Unified Agency Conditioning} to explicitly associate and model each character with its assigned roles within a unified video generation model.

\item Experiments on two shared-view 2D games demonstrate flexible character control under different role assignments and effective player-triggered reactive NPC behaviors.
\end{itemize}
\section{Related Works}

\subsection{Game World Models}

Game world models~\citep{genie3,gamefactory,matrixgame3,scope,H3-world} aim to construct interactive simulations of game environments by predicting future visual observations conditioned on player inputs. Early works such as GameNGen~\citep{GameNGen} demonstrated that diffusion models can serve as real-time neural game engines for \textit{DOOM}, while DIAMOND~\citep{diamond} showed that the visual fidelity of diffusion-based world models can substantially affect downstream policy learning. Subsequent efforts, including Matrix-Game 2.0/3.0~\citep{matrixgame2,matrixgame3}, LingBot-World~\citep{Lingbot-world,lingbot-world-v2}, GameFactory~\citep{gamefactory}, and Oasis~\citep{oasis}, have further advanced game world modeling toward streaming, long-horizon, and open-domain generation.

Despite these advances, they largely focus on player-centric interaction, where external actions directly control a predefined player character to navigate or interact with the environment. However, interactive games involve not only player control but also reactive interactions between players and NPCs. Modeling such interactions requires the world model to capture how NPC behaviors respond to the actions of players, motivating recent efforts on interactive behavior modeling.

\subsection{Interactive Behavior Modeling}

Recent works have begun to model reactive interactions between players and NPCs. Incantation~\citep{Incantation}, WanToFight~\citep{WanToFight}, and WorldMind~\citep{WorldMind}, for example, enable interactions among multiple characters by associating external actions and interactive behaviors with predefined character roles. These approaches substantially extend game world models beyond player-centric generation and enable richer character interactions.

However, their interaction structures are typically role-specific: control signals and reactive behaviors are tied to predefined characters or roles during training. Consequently, the same character cannot be readily reassigned between controlled and reactive roles across different rollouts. This limits their flexibility in multi-character games, where both the controlled character and the interaction may vary. 
Although ActionParty~\citep{actionparty} enables external control to be bound to a selected character, it does not explicitly model reactive interactions with NPCs. In contrast, our work focuses on configurable multi-character interactions, where control roles can be flexibly assigned across characters and reactive NPC behaviors can be conditioned on players' interactions.

\section{Method}\label{sec:method}
\method (Figure~\ref{fig:method}) assigns flexible control modes at initialization and jointly simulates externally controlled players and reactive NPCs. We first formulate selective agency in multi-character games, then present \textit{Spatial Role Binding} and \textit{Unified Agency Conditioning}.

\subsection{Problem Formulation}\label{sec:problem_formulation}

In a multi-character game, external actions and reactive behavior instructions must be associated with the characters that execute them. Our goal is to make them explicit: users select which characters receive external control and assign reactive behaviors to the remaining NPCs. This is particularly challenge in shared-view 2D games, where characters can share the same appearance and viewpoint cues cannot distinguish their control roles.

Given an initial observation $X_0$ containing $N$ characters, let $\mathbf{M}_0=(M_1,\ldots,M_N)$ denote their initial-frame masks. At rollout initialization, we assign each character a control mode $\rho_i\in\{\mathrm{external},\mathrm{NPC}\}$, collected in $\boldsymbol{\rho}$. An externally controlled character receives an action sequence $\mathbf{a}_i=(a_{i,0},\ldots,a_{i,T-1})$ over $T$ control intervals. Each command uses the text format \texttt{ACTION=<action>}, e.g., \texttt{ACTION=jump} or \texttt{ACTION=move\_left + jump}. An NPC receives a conditional rule $r_i$, with expressed as \texttt{If [condition], then [response]}, e.g., \texttt{If a player throws a shuriken, then jump to dodge it.} The indexed control inputs are
\begin{equation}
    \mathcal{A}=\{(i,\mathbf{a}_i):\rho_i=\mathrm{external}\},
    \qquad
    \mathcal{R}=\{(i,r_i):\rho_i=\mathrm{NPC}\}.
    \label{eq:agency-inputs}
\end{equation}
The model jointly generates the future video according to
\begin{equation}
    \mathbf{X}_{1:F}\sim p_{\theta}\!\left(
        \cdot\mid X_0,\mathbf{M}_0,\boldsymbol{\rho},\mathcal{A},\mathcal{R}
    \right),
    \label{eq:selective-agency-formulation}
\end{equation}
where $F$ is the number of output video frames. All characters evolve within the same generated scene, so their motions and interactions must remain jointly consistent.

For conditional NPC rule, it specifies a trigger condition and a response. The initial mask identifies the NPC that executes the response, while any externally controlled player may satisfy the trigger condition. For example, it may instruct an NPC to jump in response to a fireball in SF3 game. The initial mask identifies the character that should execute the response. The model must infer when the condition is met from the evolving video context and realize the response in the generated frames. Control modes and rules remain fixed within a rollout and can be reassigned when initializing another rollout.

This formulation combines external action control with conditional NPC rule, while allowing the control mode of each character to be specified independently of its appearance. Base on this formulation, we propose our method \method. For \method, Spatial Role Binding (Section~\ref{sec:spatial_role_binding}) grounds character identities in the initial frame. Unified Agency Conditioning (Section~\ref{sec:unified_agency_conditioning}) binds actions and rules to those identities and integrates them with causal video context.

% \subsection{Data Construction}\label{sec:data_construction}

% We construct datasets in two game environments, \textit{Street Fighter III}~(SF3)~\citep{sf3} and \textit{Hikou no mizu}~(HNM)~\citep{hnm}, by collecting trajectories with explicit character-level control assignments. 
% SF3 contain two fighters with up to one NPC, while HNM contain two to six characters with up to two NPCs. Each trajectory spans five seconds, comprising 101 RGB frames at 20 fps and 25 control intervals. We record the initial image, per-character initial masks, control modes, external action sequences, NPC behavior specifications, and the corresponding video.

% Externally controlled characters follow scripted sequences from the game's discrete action vocabulary, while rule-based controllers generate NPC responses from observed game states and events. We vary character appearances, initial layouts, control assignments, and action sequences. Reactive rule behaviors include approaching or retreating, evading or blocking attacks, and counterattacking, with coverage of repeated responses and passive NPCs. The details of dataset are in Appendix~\ref{apen:datasets}.

\subsection{Spatial Role Binding}\label{sec:spatial_role_binding}
\begin{figure}[!t]
    \centering
    \includegraphics[width=\linewidth]{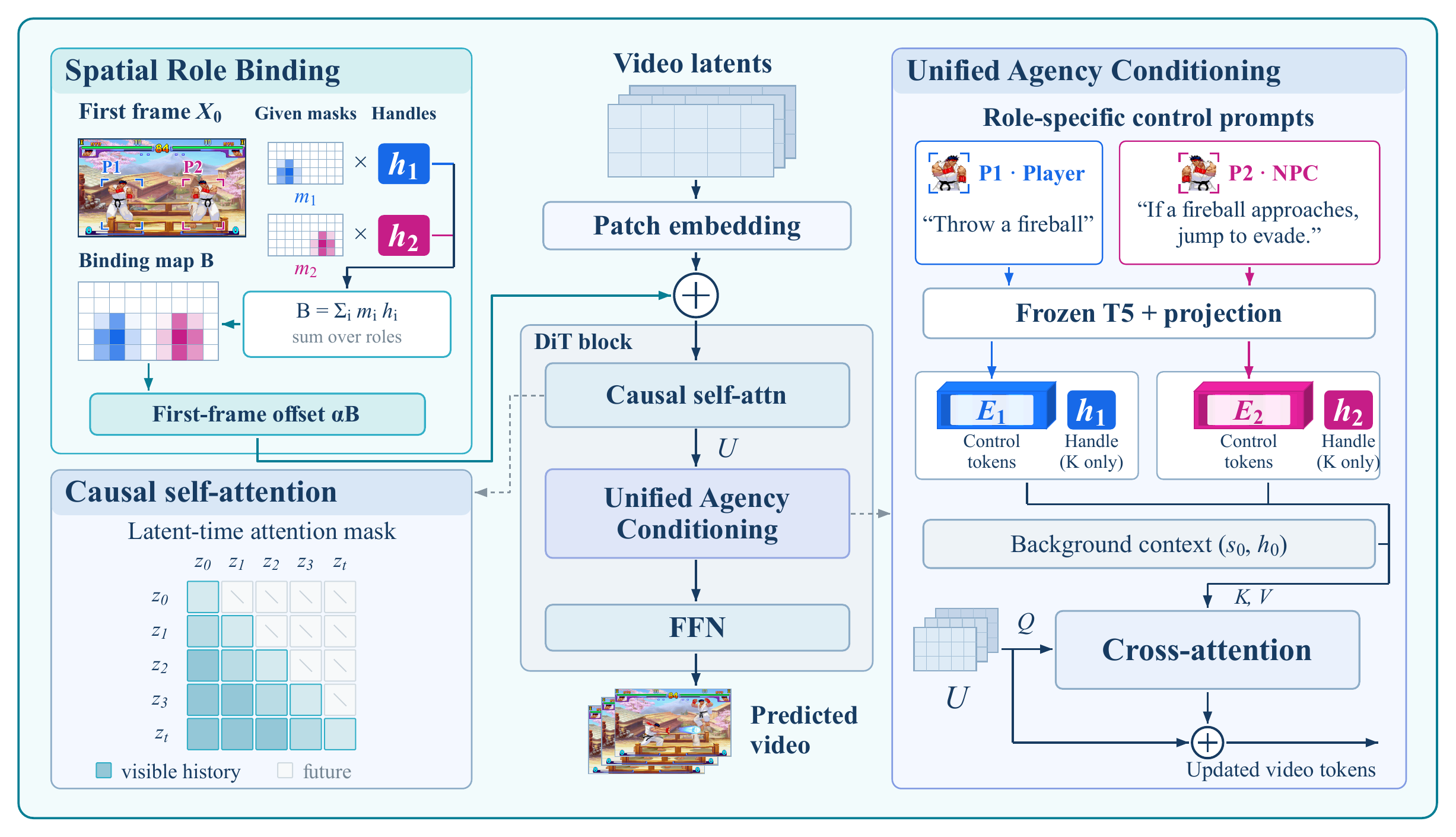}
    \caption{Overview of \method. Spatial Role Binding injects mask-weighted role handles into the first-frame tokens. Unified Agency Conditioning incorporates role-specific control tokens and background context through cross-attention, with actor handles injected into keys only. Causal self-attention limits temporal access to the current and past latent frames. }
    \label{fig:method}
\end{figure}
To realize \method, the primary challenge is establishing reliable character binding. In a shared-view 2D game, visual appearance and camera perspective are insufficient to uniquely identify the targeted character---an ambiguity compounded when multiple entities share an identical appearance. We address this by utilizing initial-frame segmentation masks to map each character instance to a learned handle $h_i \in \mathbb{R}^{d}$, as shown in Figure~\ref{fig:method}. These handles are drawn from a shared codebook and normalized to unit root-mean-square magnitude. Each handle uniquely tracks an instance throughout a rollout and its index implies neither character class nor control modality (e.g., player vs.~NPC). Finally, $h_0$ is reserved for the background.

Let $H_t(p)\in\mathbb{R}^{d}$ denote the patch embedding at spatial position $p$ in latent frame $t$, with $t=0$ corresponding to the initial frame. We resize each binary character mask to the patch grid by area averaging, obtaining soft coverage $\bar m_i(p)\in[0,1]$. To prevent overlapping masks from increasing the total binding magnitude, we use
\begin{equation}
    m_i(p)=\frac{\bar m_i(p)}{\max\!\left(1,\sum_{j=1}^{N}\bar m_j(p)\right)}.
    \label{eq:mask-coverage}
\end{equation}
The spatial binding map and the resulting patch embeddings are
\begin{equation}
    B(p)=\sum_{i=1}^{N}m_i(p)h_i,
    \qquad
    \widetilde H_t(p)=H_t(p)+\mathbf{1}[t=0]\,\alpha B(p),
    \label{eq:spatial-role-binding}
\end{equation}
where $\alpha=0.25$ controls the injection strength. Thus, the same handle used to address a character in the control pathway is introduced at that character's initial spatial location. Patches outside the character masks receive zero offset. The binding is applied only to the initial frame, so generation requires no future character masks.

During training, we permute the assignment of characters to handle slots, jointly reordering their masks, control modes, action sequences, and NPC rules. This discourages fixed associations between handle indices, appearances, and control modes. The resulting interface allows a character to receive either external actions or an NPC rule while retaining the same spatial binding mechanism.

\subsection{Unified Agency Conditioning}\label{sec:unified_agency_conditioning}

Building upon character binding, the subsequent challenge is unifying heterogeneous control signals into a single model while faithfully executing the designated behavior for each character. To this end, we propose Unified Agency Conditioning (UAC). UAC structures both external actions and reactive NPC rules as control-token groups addressed by character handles, and injects them into causal video features via a shared cross-attention mechanism. 

\paragraph{Control tokens and cross-attention.}
For character $i$ and control interval $\tau$, let $c_{i,\tau}$ denote its control text. For an externally controlled character, $c_{i,\tau}$ is the action text for that interval. For an NPC, $c_{i,\tau}=r_i$ is a conditional rule describing its trigger and response, reused across control intervals. A frozen UMT5~\citep{T5} encoder $\mathcal{T}$ and a shared trainable projection $\mathcal{P}$~\citep{Wan} produce
\begin{equation}
    E_{i,\tau}=\mathcal{P}\!\left(\mathcal{T}(c_{i,\tau})\right).
    \label{eq:control-encoding}
\end{equation}
We preserve all valid text tokens and mask padding. The NPC rule is encoded jointly, allowing its condition and response to be interpreted together from video context.

We use the projected text sequence directly as the character's control group, $C_{i,\tau}=E_{i,\tau}$. The actor handle $h_i$ binds this group to the character that executes the instruction through the cross-attention keys. We also include a background group $C_{0,\tau}=\{s_0\}$, where $s_0$ is a learnable background token shared across control intervals and paired with the background handle $h_0$.

For clarity, we suppress the block index. Let $U_t(p)$ denote the video features after self-attention and let $c_{i,\tau,\ell}$ be token $\ell$ in $C_{i,\tau}$. Queries are obtained from the normalized video features, while keys combine control content with the actor handle:
\begin{equation}
    \begin{aligned}
        q_{t,p} &= \operatorname{RMSNorm}\!\left(W_Q\operatorname{LN}(U_t(p))\right),\\
        k_{i,\tau,\ell} &= \operatorname{RMSNorm}\!\left(W_Kc_{i,\tau,\ell}+W_hh_i\right),\\
        v_{i,\tau,\ell} &= W_Vc_{i,\tau,\ell}.
    \end{aligned}
    \label{eq:handle-bound-kv}
\end{equation}
The projection notation includes affine biases where present. RMS normalization precedes the split into attention heads; in the attention below, $q$, $k$, and $v$ refer to the corresponding per-head slices. The actor handle is broadcast to all keys in its group, allowing queries to select controls by both their content and their bound character. It is injected only into keys and values are obtained from the projected control text.

Let $L_{i,\tau}$ denote the number of valid tokens in control group $i$. To account for different group lengths, we normalize the attention logits as
\begin{equation}
    a_{t,p,i,\ell}
    =\operatorname*{softmax}_{j,m}\!\left(
        \frac{q_{t,p}^{\top}k_{j,\tau,m}}{\sqrt{d_h}}
        -\log L_{j,\tau}+\beta_j
    \right)_{i,\ell},
    \qquad \tau=t-1,
    \label{eq:group-attention}
\end{equation}
where $d_h$ is the head dimension and the softmax ranges over all valid tokens in all groups, including the background. For each future latent frame $t\in\{1,\ldots,T\}$, we use control interval $\tau=t-1$, since $t=0$ is reserved for the initial frame. The length correction prevents a group from receiving a larger baseline attention mass solely because it contains more tokens. We set $\beta_i=0$ for character groups and $\beta_0=\log N$ for the background, balancing its baseline mass against the collection of $N$ character groups. The background value is multiplied by a learned scale initialized to zero. The head outputs $\sum_{i,\ell}a_{t,p,i,\ell}v_{i,\tau,\ell}$ are concatenated, projected, and added to $U_t(p)$. The UAC residual is zero for the initial frame.

\paragraph{Causal video context.}
Reactive behavior requires access to the evolving scene. Before the cross-attention update, self-attention aggregates video context using a mask defined over latent-frame indices:
\begin{equation}
    \mathcal{M}^{\mathrm{causal}}_{(t,p),(s,q)}=
    \begin{cases}
        0, & s\leq t,\\
        -\infty, & s>t.
    \end{cases}
    \label{eq:causal-video-mask}
\end{equation}
Consequently, every spatial token in frame $t$ can attend to all positions in its current frame and preceding frames, including the initial spatial bindings. The temporal restriction applies between frames; spatial attention within a frame remains unrestricted. UAC uses the resulting features $U_t(p)$ to interpret an NPC rule in the context of the visible interaction. Control interval $\tau$ is aligned with future latent frame $t=\tau+1$, and cross-attention at that frame reads only the corresponding control groups. Besides, we train the model with standard latent flow matching over future video latents, keeping the initial-frame latent clean. Together, these restrictions prevent future video tokens or later external controls from entering an earlier frame's conditioning pathway.

\section{Experiments}\label{sec:experiments}

\subsection{Experimental Settings}\label{sec:exp_settings}
\paragraph{Datasets.} We utilize two 2D-game environments: Hikou no Mizu (HNM)~\citep{hnm} and Street Fighter III: New Generation (SF3)~\citep{sf3}. Both protocols use five-second clips containing 101 frames at 20 FPS and $832 \times 480$ resolution. For HNM, it comprising 20,000 training clips with 2--6 characters and up to two NPCs. It combines direct action control with nine trigger--response rules and a no-op category. For SF3, it covers two-fighter interactions across eleven playable characters, targeting 20,000 training clips with NPC interactions covering nine reaction rules and a no-op category. The details are illustrated in Appendix~\ref{apen:datasets}.

\textbf{Models.} We initialize \method from the pretrained Wan2.2-TI2V-5B model~\citep{Wan} and train separate models for HNM and SF3. The diffusion transformer and the proposed role-binding and conditioning modules are jointly fine-tuned, while the UMT5-XXL text encoder and Wan2.2 VAE remain frozen. We use AdamW with a learning rate of $5\times10^{-5}$, weight decay $0.01$, and $(\beta_1,\beta_2)=(0.9,0.999)$. Training uses BF16 precision and an effective batch size of 8 for both HNM and SF3, with each trained for 20K steps. We optimize the weighted flow-matching MSE over the 25 future latent frames, keeping the initial-frame latent clean and excluding it from the loss.

\textbf{Evaluation Metrics.} Action accuracy (AA) measures correct action execution by the instructed player without unrelated active behavior, and selection accuracy (SA) measures whether the selected player jumps while others remain idle. 
Joint reactivity (JR) measures reference-aligned waiting and response onset: the NPC must remain inactive before a fixed trigger boundary from the paired GT trajectory and initiate the prescribed response afterward.
Note that the no-op cases require inactivity throughout the clip. These three metrics are case-level pass rates. Visual quality (VQ) equally weights clarity, character identity and structure, scene integrity, and temporal stability. Using the evaluation data collected for JR, VQ assesses the visual fidelity of generated player–NPC interactions. All scores use a 0--100 scale. We report mean $\pm$ sample standard deviation over three rounds separately for GPT-6-astra~\citep{gpt6} and Gemini 3.5 Flash~\citep{gemini}. The details are illustrated in Appendix~\ref{apen:evaluation}.

\subsection{Main Results}\label{sec:main_result}

\begin{table}[htbp]
    \centering
    \caption{Comparison of model variants under Gemini 3.5 Flash (Gemini) and GPT-6-astra (GPT) evaluation for AA, SA, $\mathrm{JR}$, and VQ. GT serves as an empirical upper-bound reference.}
    \label{tab:model-variant-comparison}
    \small
    \setlength{\tabcolsep}{4pt}
    \renewcommand{\arraystretch}{1.2}
    \resizebox{\linewidth}{!}{%
    \begin{tabular}{l*{8}{c}}
        \toprule
        & \multicolumn{2}{c}{\textbf{AA} $\uparrow$}
        & \multicolumn{2}{c}{\textbf{SA} $\uparrow$}
        & \multicolumn{2}{c}{\textbf{JR} $\uparrow$}
        & \multicolumn{2}{c}{\textbf{VQ} $\uparrow$} \\
        \cmidrule(lr){2-3}\cmidrule(lr){4-5}\cmidrule(lr){6-7}\cmidrule(lr){8-9}
        & Gemini & GPT & Gemini & GPT & Gemini & GPT & Gemini & GPT \\
        \midrule
        \rowcolor{black!7}
        HNM-GT & 97.53$\pm$2.14 & 100.00$\pm$0.00 & 97.78$\pm$0.96 & 100.00$\pm$0.00 & 98.89$\pm$1.92 & 100.00$\pm$0.00 & 98.69$\pm$0.18 & 100.00$\pm$0.00 \\
        \textbf{\method} & 88.89$\pm$0.00 & 96.30$\pm$0.00 & 95.00$\pm$4.41 & 100.00$\pm$0.00 & 76.67$\pm$0.00 & 76.67$\pm$0.00 & 79.31$\pm$0.37 & 78.80$\pm$0.53 \\
        \bottomrule
        \rowcolor{black!7}
        SF3-GT & 100.00$\pm$0.00 & 96.97$\pm$0.00 & 100.00$\pm$0.00 & 100.00$\pm$0.00 & 96.67$\pm$3.33 & 100.00$\pm$0.00 & 99.65$\pm$0.12 & 99.79$\pm$0.00 \\
        \textbf{\method} & 98.99$\pm$1.75 & 96.97$\pm$0.00 & 100.00$\pm$0.00 & 100.00$\pm$0.00 & 91.11$\pm$1.92 & 91.11$\pm$3.85 & 71.85$\pm$1.27 & 82.49$\pm$0.52 \\
        \bottomrule
    \end{tabular}%
    }
\end{table}

\begin{figure}[H]
    \centering
    \includegraphics[width=0.95\linewidth]{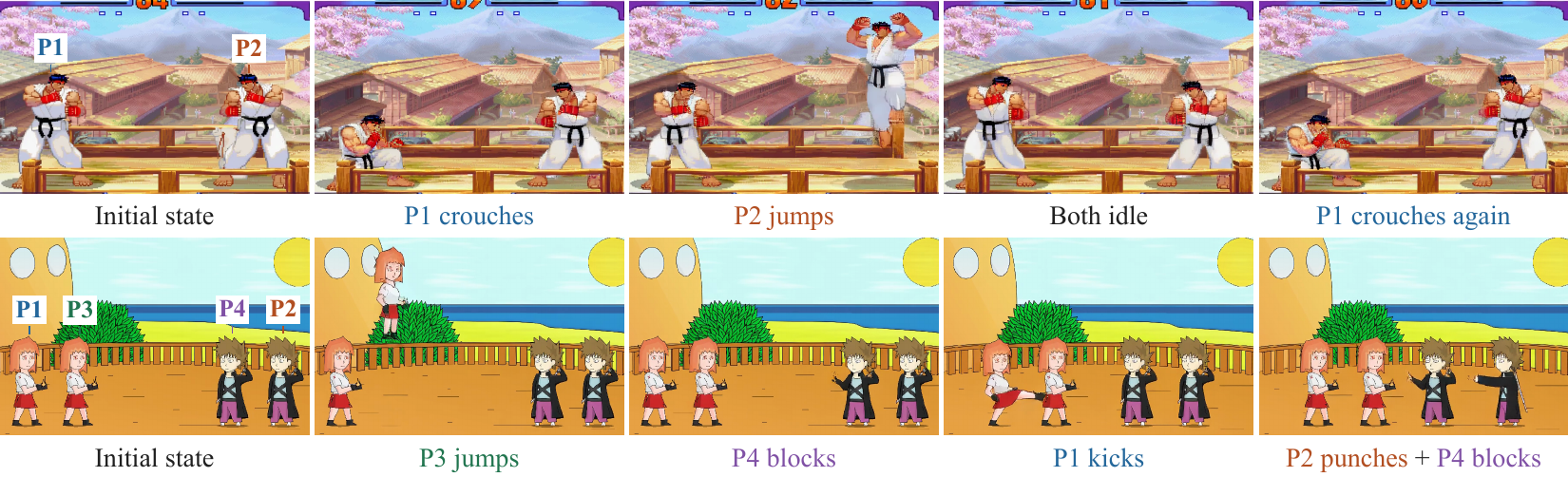}
    \caption{Multi-player interaction generated by \method in SF3 (top) and HNM (bottom). Colors match players to their actions.}
    \label{fig:multiplayer-interaction}
\end{figure}
\paragraph{Character selection and action execution.} This paragraph evaluates the performance of \method in character selection and player action injection.
Table~\ref{tab:model-variant-comparison} compares \method with GT under both judges. On SF3, AA reaches 98.99/96.97 (Gemini/GPT), close to the GT scores of 100.00/96.97, while HNM AA is 88.89/96.30 compared with 97.53/100.00 for GT. \method achieves SA of 95.00/100.00 on HNM and 100.00/100.00 on the SF3 selection cases. These results demonstrate that \method reliably selects the intended character in both environments while effectively incorporating player actions into the model.

Figure~\ref{fig:multiplayer-interaction} complements these results by illustrating independent and simultaneous character control. In SF3, two identically dressed fighters separately execute P1's crouch and P2's jump, then return to idle before P1 crouches again. In HNM, four characters, including pairs with the same appearance, execute different action sequences. The final frame shows P2 punching while P4 blocks under simultaneous inputs. These examples illustrate both separate control of visually similar characters and concurrent execution of different actions. 

\begin{figure}[H]
    \centering
    \includegraphics[width=0.95\linewidth]{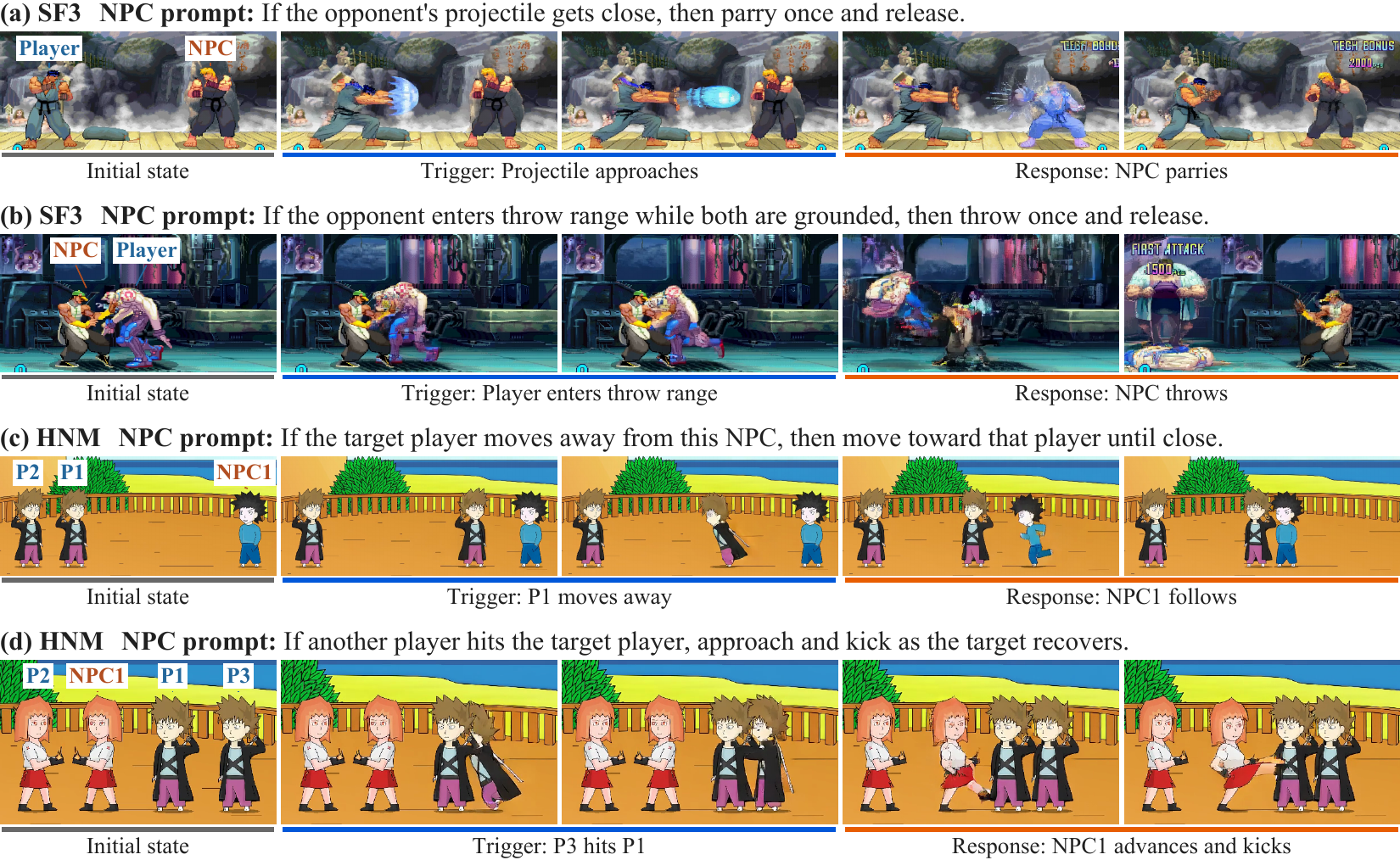}
    \caption{NPC reactivity in SF3 (a,b) and HNM (c,d). Each generated sequence shows the initial roles, a player-triggered event, and the NPC response. }
    \label{fig:npc-reactivity}
\end{figure}

\paragraph{NPC reactivity and visual quality.} This paragraph evaluates \method in terms of NPC reactivity and visual quality in reactive game generation.
\method achieves $\mathrm{JR}$ scores of 76.67/76.67 on HNM and 91.11/91.11 on SF3 (Gemini/GPT). 
Under this reference-aligned protocol, both judges assign higher JR scores to SF3 than to HNM, although scores remain below the GT references of 98.89/100.00 on HNM and 96.67/100.00 on SF3. 

Figure~\ref{fig:npc-reactivity} illustrates the requested trigger--response reactivity in selected generated sequences. The SF3 examples show a projectile parry and a throw triggered when the player enters close range. In HNM, one example depicts an NPC following a departing player, while another shows an NPC advancing and kicking after a third character strikes the player. The latter demonstrates that NPC responses can be conditioned on interactions between other characters, extending beyond direct player--NPC exchanges. More reactivity results are provided in Appendix~\ref{apen:all_reactivity}, while additional results on multi-player interactions are presented in Appendix~\ref{apen:multiple_npcs}. VQ scores are 79.31/78.80 on HNM and 71.85/82.49 on SF3, below the corresponding GT scores of 98.69/100.00 and 99.65/99.79. Thus, despite strong action control and reactive behavior, visual fidelity remains an area for improvement.

\begin{table}[H]
    \centering
    \caption{Component ablations of \method on HNM game.}
    \label{tab:hnm-ablation}
    \small
    \setlength{\tabcolsep}{2pt}
    \renewcommand{\arraystretch}{1.2}
    \resizebox{\linewidth}{!}{%
    \begin{tabular}{l*{8}{c}}
        \toprule
        & \multicolumn{2}{c}{\textbf{AA} $\uparrow$}
        & \multicolumn{2}{c}{\textbf{SA} $\uparrow$}
        & \multicolumn{2}{c}{\textbf{JR} $\uparrow$}
        & \multicolumn{2}{c}{\textbf{VQ} $\uparrow$} \\
        \cmidrule(lr){2-3}\cmidrule(lr){4-5}\cmidrule(lr){6-7}\cmidrule(lr){8-9}
        & Gemini & GPT & Gemini & GPT & Gemini & GPT & Gemini & GPT \\
        \midrule
        \rowcolor{black!7}
        HNM-GT & 97.53$\pm$2.14 & 100.00$\pm$0.00 & 97.78$\pm$0.96 & 100.00$\pm$0.00 & 98.89$\pm$1.92 & 100.00$\pm$0.00 & 98.69$\pm$0.18 & 100.00$\pm$0.00 \\
        \textbf{\method} & 88.89$\pm$0.00 & 96.30$\pm$0.00 & 95.00$\pm$4.41 & 100.00$\pm$0.00 & 76.67$\pm$0.00 & 76.67$\pm$0.00 & 79.31$\pm$0.37 & 78.80$\pm$0.53 \\
        \midrule
         w/o Spatial Role Binding & 66.67$\pm$0.00 & 54.32$\pm$2.14 & 15.00$\pm$0.00 & 15.00$\pm$0.00 & 26.67$\pm$0.00 & 23.33$\pm$0.00 & 79.39$\pm$0.83 & 78.24$\pm$0.18 \\
         w/o NPC Conditioning & 88.89$\pm$0.00 & 87.65$\pm$2.14 & 95.56$\pm$0.96 & 98.33$\pm$0.00 & 17.78$\pm$5.09 & 13.33$\pm$0.00 & 60.30$\pm$0.65 & 63.12$\pm$0.18 \\
        %w/o Causal Self-Attention & --- & --- & --- & --- & --- & --- & --- & --- \\
        \bottomrule
    \end{tabular}%
    }
\end{table}

\subsection{Effectiveness of Spatial Role Binding}\label{sec:effectiveness_of_spatial}

In this section, We investigate whether \textit{Spatial Role Binding} helps associate video features with the correct character's controls and how it establishes this correspondence.

We first compare the full model with the no-binding variant on HNM using AA, SA, $\mathrm{JR}$, and VQ. Table~\ref{tab:hnm-ablation} shows that removing Spatial Role Binding reduces AA from 88.89/96.30 to 66.67/54.32, SA from 95.00/100.00 to 15.00/15.00, and $\mathrm{JR}$ from 76.67/76.67 to 26.67/23.33 (Gemini/GPT). In contrast, VQ remains similar, at 79.31/78.80 for the full model and 79.39/78.24 for the no-binding variant. These results suggest that Spatial Role Binding primarily benefits character-specific control and NPC reactivity, with little change in overall visual quality.

To investigate the mechanism underlying these gains, we analyze layer-wise cross-attention on 200 HNM validation videos with noised GT latents at noise fractions of 0.1, 0.5, and 0.9. At each layer, we sum attention probabilities over each character's control tokens and average across heads. Foreground actor-selection accuracy measures whether the highest-scoring character group matches the query's character, with the background group excluded from Top-1 selection.

\begin{figure}[H]
    \centering
    \includegraphics[width=\linewidth]{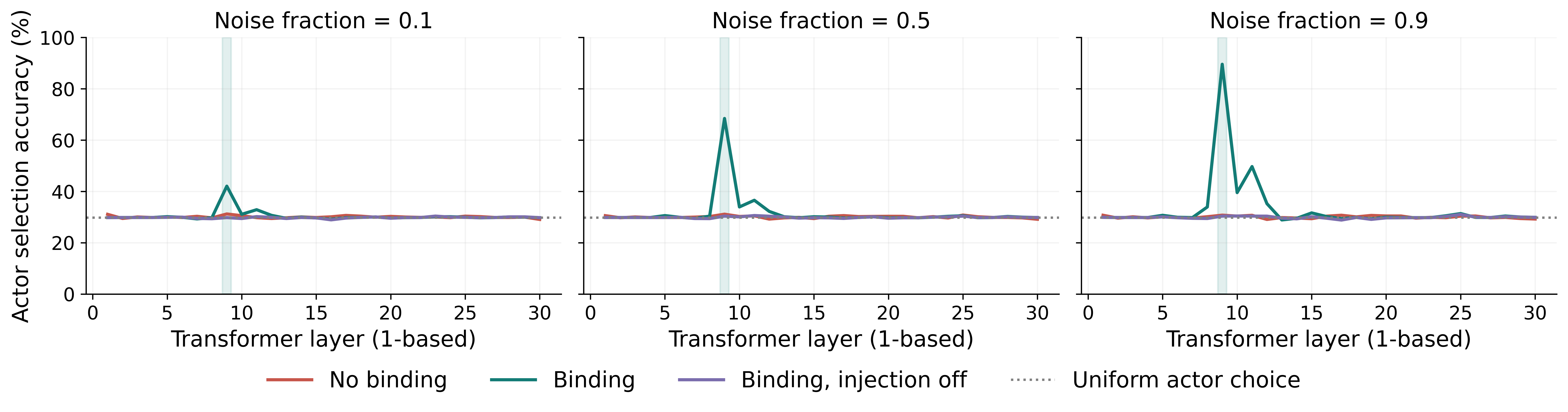}
    \caption{Layer-wise foreground actor-selection accuracy on 200 HNM validation videos with noised GT latents. Attention probabilities are summed by actor and averaged across heads; Top-1 excludes the background group. The dotted line denotes uniform actor choice.}
    \label{fig:binding-layer-accuracy}
\end{figure}

We compare \method, the no-binding variant, and \method with first-frame handle injection disabled. Figure~\ref{fig:binding-layer-accuracy} shows a prominent layer-9 peak with binding, most pronounced at high noise. Disabling first-frame injection removes this peak and brings accuracy close to uniform actor choice.
Together, these findings support Spatial Role Binding as a spatial cue that links video queries to the appropriate character's control group.

\subsection{Analysis on NPC Conditioning}\label{sec:comparison_different_condition}

\begin{wrapfigure}{r}{0.48\textwidth}
    \vspace{-\intextsep}
    \captionsetup{font=small,skip=4pt}
    \centering
    \includegraphics[width=\linewidth]{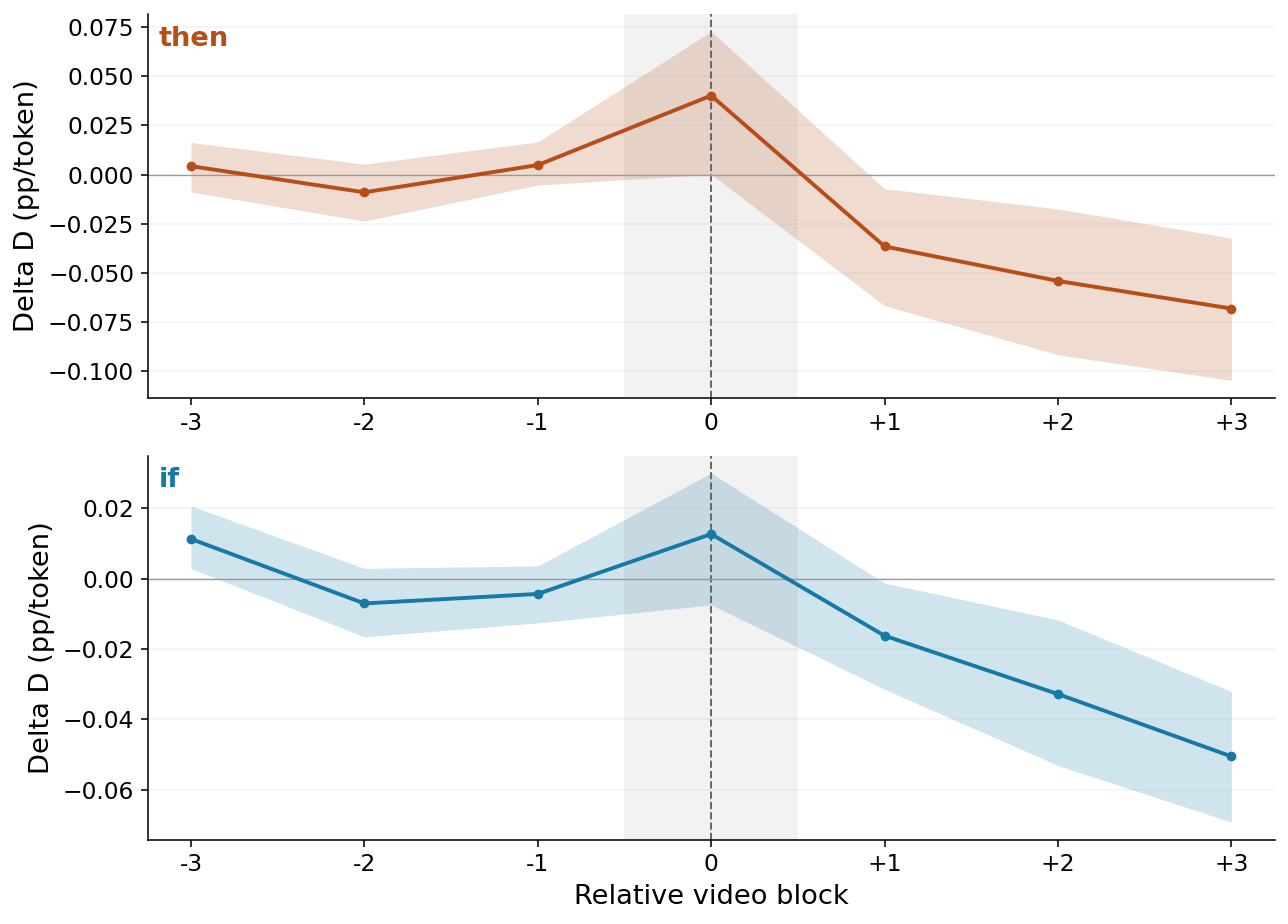}
    \caption{Layer-9 attention to NPC response (\textit{then}, top) and condition (\textit{if}, bottom) clauses. Curves show changes in attention per token relative to the three-block pre-trigger baseline, averaged with equal rule weights. Shading denotes 95\% initial-frame-group cluster-bootstrap intervals. Block 0 contains the source GT trigger.}
    \label{fig:npc-conditioning-attention}
\end{wrapfigure}
\vspace{-1em}

We investigate whether NPC conditioning supports rule-guided reactions and how NPC video features attend to the condition and response clauses during generation.
We first compare the full model with the variant without NPC conditioning on HNM. Table~\ref{tab:hnm-ablation} shows that removing NPC conditioning reduces $\mathrm{JR}$ from 76.67/76.67 to 17.78/13.33 and VQ from 79.31/78.80 to 60.30/63.12 (Gemini/GPT). AA changes from 88.89/96.30 to 88.89/87.65, while SA remains high at 95.56/98.33 compared with 95.00/100.00 for the full model. These results suggest that NPC conditioning is particularly important for trigger--response behavior and also benefits the visual quality of generated player--NPC interactions.

To examine attention to the these conditional text, we conduct a separate diagnostic analysis on HNM inputs. We analyze layer 9, selected from the preceding role-binding study. During free generation, we measure attention from generated NPC body regions to their own \textit{if} and \textit{then} clauses, averaging over 24 heads and 30 denoising steps. Attention is normalized over all valid conditioning keys. For each clause, we sum its token probabilities, divide by its token count, and subtract the mean over three complete pre-trigger blocks. We align these changes to the source GT trigger block (0.2 seconds); GT timestamps and future NPC actions are not supplied to the model.

Figure~\ref{fig:npc-conditioning-attention} shows the test results, with rules weighted equally. At the GT block, 58 measurable events (trigger-response events) yield a mean \textit{then} increase of $0.04017$ percentage points per token. The lower bound is close to zero, and the increase does not persist over subsequent blocks. The \textit{if} change is $0.01271$, providing no evidence of a consistent decrease. Together, these findings show the effectiveness of conditional NPC rule for reactive NPCs generation.

% \subsection{Analysis on Casual Self-Attention}\label{sec:analysis_on_causal}

\section{Conclusions}
We presented \method, a reactive game world model for flexible role control and NPC reactivity. \textit{Spatial Role Binding} grounds character handles in the initial frame, while \textit{Unified Agency Conditioning} binds player actions and conditional NPC rules to corresponding characters. Results on HNM and SF3 demonstrate flexible role assignment and player-triggered NPC responses, supporting character-level conditioning for configurable multi-character interaction. 

\section{Acknowledgment}
We would like to express our sincere gratitude to Ruidong Wang and Murphy Zhao for their tremendous support throughout this project. We are also deeply thankful to Shusen Wang for his invaluable assistance with technical maintenance.

\bibliography{iclr2027_conference}
\bibliographystyle{iclr2027_conference}

\newpage
\appendix
\section{Details of Datasets}\label{apen:datasets}

\subsection{Shared Data Format}\label{apen:dataset_format}

Both datasets contain five-second game trajectories, represented by an initial observation and 100 future RGB frames at 20 FPS and $832\times480$ resolution. Each trajectory has 25 external-control intervals of 0.2 seconds, with each interval aligned to four future video frames. We store the initial RGB image, one initial-frame mask per character instance, control-mode assignments, external action sequences, conditional NPC rule specifications, and the target video. Character masks are obtained from native instance-ID rendering and remain associated with physical character instances even when appearances are identical. Control assignments remain fixed within a trajectory.

The model is conditioned on the initial image and masks, external actions, and static NPC rule. Realized NPC inputs, native game events, and trigger/response timestamps are retained for collection verification and are not supplied as future conditioning. Recorded target identities likewise do not introduce a separate target-slot input to \method. Table~\ref{tab:dataset-splits} summarizes the dataset splits.

\begin{table}[htbp]
    \centering
    \caption{Training, validation, and test set sizes in video clips.}
    \label{tab:dataset-splits}
    \small
    \setlength{\tabcolsep}{10pt}
    \renewcommand{\arraystretch}{1.15}
    \begin{tabular}{lrrr}
        \toprule
        Dataset & Train & Validation & Test \\
        \midrule
        HNM & 20,000 & 200 & 200 \\
        SF3 & 20,000 & 200 & 200 \\
        \bottomrule
    \end{tabular}
\end{table}

\subsection{Hikou no Mizu}\label{apen:hnm_dataset}

\paragraph{Scene composition and actions.}
HNM trajectories contain two to six characters, with zero, one, or two NPCs. The training set contains 20,000 clips covering external-only control and NPC interactions; each validation and test split contains 100 clips of each type. We vary character appearances, initial layouts, action sequences, and role assignments. The external action vocabulary has 21 commands: no-op; jump, punch, kick, and shuriken throw; crouch alone or combined with punch, kick, or shuriken throw; block alone or combined with crouch; and left/right movement alone or combined with jump, punch, kick, or shuriken throw.

\paragraph{NPC Reactive behaviors.}
Table~\ref{tab:hnm-rule-catalog} lists nine trigger--response rule families (H01--H09) and one passive no-op condition (H10). The missed-attack and jump-in rules each have separate punch and kick policies, yielding 11 active rule/policy combinations. The no-op condition releases all controls throughout the clip. NPC instructions express the trigger and response in conditional language, for example, \textit{``If the target player throws a shuriken, then jump to dodge it.''} Native controllers verify the triggering event, the corresponding response, and recovery before a subsequent response. The collection includes repeated reactions, passive NPCs, and cases in which the specified trigger does not occur.

\begin{table}[htbp]
    \centering
    \caption{HNM NPC rule families and the no-op condition. The target is the external player designated by the collection controller. These descriptions summarize the rule semantics.  Although the original prompt templates use the phrase ``the target player,'' this phrase is not accompanied by a player identifier or a target-selection input.}
    \label{tab:hnm-rule-catalog}
    \small
    \setlength{\tabcolsep}{4pt}
    \renewcommand{\arraystretch}{1.2}
    \begin{tabular}{p{0.06\linewidth}p{0.40\linewidth}p{0.45\linewidth}}
        \toprule
        Rule & Trigger & Response \\
        \midrule
        H01 & Target moves away. & Approach the target until close. \\
        H02 & Target approaches and enters close range. & Retreat until at a safe distance. \\
        H03 & Target's punch or kick misses. & Approach and punch or kick during the target's recovery. \\
        H04 & Target throws a shuriken. & Jump to dodge the projectile. \\
        H05 & Another external player hits the target. & Approach and kick as the target recovers. \\
        H06 & Target hits the NPC. & Recover, then approach and kick once. \\
        H07 & Target punches or kicks the NPC. & Block standing for high attacks and crouching for low attacks. \\
        H08 & Target jumps toward the NPC. & Punch or kick the target while airborne. \\
        H09 & Target's punch or kick contacts another external player. & Approach the target and kick once. \\
        H10 & None (no-op). & Keep all controls released throughout the clip. \\
        \bottomrule
    \end{tabular}
\end{table}

\paragraph{Training-set filtering.}
The training set includes external-only clips, passive-NPC clips, and clips with valid active reactions. For a training clip containing active NPCs, every active NPC must have a valid trigger and complete its observed trigger--response--recovery cycles, with controls released outside the response. If any active NPC has no valid trigger, the entire clip is excluded. Retained clips preserve their full waiting and recovery segments.

\subsection{Street Fighter III: New Generation}\label{apen:sf3_dataset}

\paragraph{Scene composition and roles.}
SF3 trajectories contain two fighters drawn from eleven playable characters, with either two externally controlled players or one player and one NPC. The released training set contains 10,000 clips of each type; validation and test each contain 100 clips of each type. Within the NPC subset, each of the nine reactive rules and the no-op category contributes 1,000 training clips and ten clips to each held-out split. For each rule and split, the NPC occupies each physical player slot equally often. This assignment follows the physical controller port and character instance throughout the clip, including when the fighters cross on screen.

\paragraph{Actions and temporal alignment.}
The player action vocabulary contains 20 commands: no-op, left/right movement, jump, crouch, three strengths each of punch and kick, left/right movement combined with jump, left/right movement with a light punch, crouching heavy punch, crouching light/heavy kick, and left/right crouching inputs. NPC collection uses a separate 18-command vocabulary covering movement, jump, guards, parries, attacks, and throws; these realized NPC commands are used for replay and auditing rather than model conditioning.

\paragraph{Reactive behaviors.}
Table~\ref{tab:sf3-rule-catalog} summarizes the nine conditional NPC rules (R01--R09) and the no-op category (R10). The collection includes valid zero-trigger cases, single responses, and repeated responses where supported by the rule and character configuration. Zero-trigger cases retain the same static rule instruction and require the NPC to release all controls; a missed response to a valid trigger is a failed sample, not a zero-trigger example. The passive no-op category instead specifies released controls throughout the clip. SF3 retains its valid zero-trigger cases and does not use the HNM training-set filter.

\begin{table}[htbp]
    \centering
    \caption{SF3 NPC rule families and the no-op category. The opponent is the externally controlled fighter. Native game events are used to verify the corresponding response.}
    \label{tab:sf3-rule-catalog}
    \small
    \setlength{\tabcolsep}{4pt}
    \renewcommand{\arraystretch}{1.2}
    \begin{tabular}{p{0.06\linewidth}p{0.40\linewidth}p{0.45\linewidth}}
        \toprule
        Rule & Trigger & Response \\
        \midrule
        R01 & Opponent moves away. & Approach until close. \\
        R02 & Opponent enters close range. & Retreat to a safe distance. \\
        R03 & Opponent's attack misses. & Punish the same attack during recovery. \\
        R04 & Opponent's projectile threatens the NPC. & Jump to evade that projectile. \\
        R05 & Opponent hits the NPC. & Recover before counterattacking. \\
        R06 & Opponent attacks the NPC. & Guard according to the attack type. \\
        R07 & Opponent jumps toward the NPC. & Anti-air the opponent while airborne. \\
        R08 & Opponent's projectile approaches. & Parry the projectile. \\
        R09 & Opponent enters throw range. & Execute a normal throw. \\
        R10 & None (no-op). & Keep all controls released throughout the clip. \\
        \bottomrule
    \end{tabular}
\end{table}

\paragraph{Quality control.}
The initial RGB image matches video frame zero, and each mask contains the visible primary character body, excluding projectiles, shadows, effects, and interface elements. We verify that recorded action IDs reconstruct the complete native input sequence and that native events confirm trigger and response outcomes. Duplicate checks cover RGB content, character-ID renderings, full inputs, and character trajectories, including mirrored or role-swapped variants.

\section{Details of Evaluation}\label{apen:evaluation}

Following Section~\ref{sec:method}, $F=100$ denotes the number of future RGB frames, $T=25$ the number of control intervals, and $\tau\in\{0,\ldots,T-1\}$ a control-interval index. We use $f\in\{0,\ldots,F\}$ for RGB frames, with $f=0$ corresponding to $X_0$, and retain $t$ for latent frames. Evaluation cases, judges, and rounds are indexed by $n$, $g$, and $e\in\{1,2,3\}$, respectively. Scores are computed separately for each game and model variant.

For AA, SA, and JR, the judge returns \texttt{pass}, \texttt{fail}, or \texttt{unknown}, together with frame-indexed observations and an explanation. Both \texttt{fail} and \texttt{unknown} remain in the denominator and receive zero credit. Let $K_\mu$ be the number of cases for metric $\mu\in\{\mathrm{AA},\mathrm{SA},\mathrm{JR}\}$, and let $y_{n,g,e}^{\mu}$ denote the case-level judgment. The round score is
\begin{equation}
    S_{g,e}^{\mu}=\frac{100}{K_\mu}\sum_{n=1}^{K_\mu}\mathbf{1}\!\left[y_{n,g,e}^{\mu}=\texttt{pass}\right].
    \label{eq:eval-case-score}
\end{equation}
We first aggregate cases within each round, then report the mean and sample standard deviation across the three round scores, separately for each judge. The same procedure is used for the round-level VQ scores defined below.

\subsection{Action Accuracy (AA)}\label{apen:eval_aa}

\paragraph{Objective and data.}
AA measures whether an action instruction is visibly executed by its designated character without unrelated active behavior. HNM tests nine basic actions: left/right movement, jump, punch, kick, shuriken throw, block, crouch, and no-op. Each action has three independently constructed scene variants, giving 27 cases. Scenes contain two visually distinct characters in an open layout: one receives the tested action, while the other receives no-op. Brief actions are requested once; sustained actions occupy a short control interval surrounded by no-op intervals. SF3 tests 11 actions---no-op, left/right movement, jump, crouch, and light/medium/heavy punches and kicks---in three scene variants with Ryu, Yang, or Ken as the tested fighter, giving 33 cases. 

\paragraph{Evidence and decision.}
Each judge receives 25 full-clip frames at $f_u=\operatorname{round}(Fu/24)$ for sample index $u\in\{0,\ldots,24\}$, together with a dense action window; duplicate frame indices are removed. For HNM, an action scheduled at control interval $\tau_{\mathrm{act}}$ uses every frame from $4\tau_{\mathrm{act}}-2$ through $4\tau_{\mathrm{act}}+18$, clipped to $[0,F]$. The factor of four is the number of future RGB frames per control interval. SF3 AA uses the frozen interval 16--44 (0.8--2.2 seconds), around the scheduled 1.0-second action start. No-op receives the same sampling procedure. Each case receives one direct judgment per round. The dense window captures brief actions, while full-clip frames expose additional activity before or after the requested action.

Passing requires the designated instance to perform the requested action with no unrelated new active action in the supplied evidence. Movement requires its own displacement in the requested image direction; jumping requires takeoff; punches and kicks require the specified strike but not a hit. HNM throwing also requires a visible shuriken, while block and crouch require the corresponding posture. For SF3, an unresolved punch/kick strength is \texttt{unknown}, not credited as a generic strike. Ordinary idle animation, passive recoil, natural physics, and action recovery are allowed. The instructed action or its name is not treated as evidence that it occurred.

\subsection{Selection Accuracy (SA)}\label{apen:eval_sa}

SA isolates instance selection by using the same requested action, a single jump, for every tested role. HNM has three independent scenes for each population size from two to six. Within each scene, every character is selected once, yielding $3(2+3+4+5+6)=60$ cases. Paired selections share the initial scene, identity assignments, timing, and generation seed, changing only the selected character. This includes visually identical characters as separate instances. SF3 uses three two-fighter scenes, including an identical-character pair; selecting each fighter once yields six cases.

Both games use the 25 full-clip frames and the input-timed dense window $[4\tau_{\mathrm{act}}-2,4\tau_{\mathrm{act}}+18]$, clipped to $[0,F]$, where $\tau_{\mathrm{act}}$ is the scheduled jump interval. A case passes only if the selected physical instance jumps once and every other character performs no new active action. Selecting the wrong jumper, multiple jumpers, no jumper, or another active action fails. Natural idle motion and completion of a pre-existing animation are allowed. SA is averaged over all cases, so the HNM aggregate weights population sizes by their case counts rather than equally weighting the five population sizes. The six-case SF3 suite provides a limited selection diagnostic.

\subsection{Joint Reactivity (\texorpdfstring{$\mathrm{JR}$}{JR})}\label{apen:eval_ra}
\paragraph{Objective and cases.}
The reported $\mathrm{JR}$ tests whether the designated NPC waits before a fixed GT trigger boundary and initiates the prescribed response afterward. Each game uses 30 cases: three GT-trigger cases for each of the nine active reaction categories, plus three passive no-op cases. Active categories follow H01--H09 for HNM and R01--R09 for SF3; the no-op categories are H10 and R10, respectively. For no-op cases, waiting is required on both sides of a fixed random split.
The category semantics follow Tables~\ref{tab:hnm-rule-catalog} and~\ref{tab:sf3-rule-catalog}. Table~\ref{tab:ra-response-criteria} states the post-boundary action criteria used for evaluation.

\begin{table}[H]
    \centering
    \caption{Response actions used by the visual JR evaluator. An eligible action must begin in a scored post-boundary window. No-op instead requires continued waiting.}
    \label{tab:ra-response-criteria}
    \small
    \setlength{\tabcolsep}{5pt}
    \renewcommand{\arraystretch}{1.15}
    \begin{tabular}{lp{0.64\linewidth}}
        \toprule
        Rules & Accepted response \\
        \midrule
        HNM H01 / H02 & NPC's own movement toward / away from any external player \\
        HNM H03, H08 & Punch \\
        HNM H04 & Jump \\
        HNM H05, H06, H09 & Kick \\
        HNM H07 & Standing or crouching block \\
        SF3 R01 / R02 & NPC's own movement toward / away from the other fighter \\
        SF3 R03, R05, R07 & Attack, without requiring a particular strike strength \\
        SF3 R04 / R06 & Jump / standing or crouching block \\
        SF3 R08 / R09 & Parry / normal throw \\
        HNM H10, SF3 R10 & Wait throughout the scored segment \\
        \bottomrule
    \end{tabular}
\end{table}

\paragraph{Fixed temporal alignment.}
For each active case, let $f_n^{\mathrm{bdry}}$ denote the RGB frame corresponding to the trigger timestamp in its native GT event log. This boundary is fixed for both GT and generated videos and is not adjusted to events in the generated clip. Frames $1,\ldots,f_n^{\mathrm{bdry}}-1$ form the waiting segment, and frames $f_n^{\mathrm{bdry}},\ldots,F$ form the response segment. No-op cases use a fixed random split. GT event times and realized NPC actions are not provided to the model.

\paragraph{Window judgments and aggregation.}
JR covers all 100 future frames. Each segment is partitioned into consecutive windows of at most 44 scored frames, preceded by up to five context frames, plus the identity reference. Context helps identify ongoing actions and new onsets but is not scored again. A WAIT window passes only when the NPC shows no deliberate activity. New movement, jumps, attacks, guard, crouch, parry, or throws violate waiting; ordinary idle animation, passive recoil, and natural physics do not. An ongoing deliberate action still violates waiting even if it began in context.

A RESPOND window passes only if the required action newly starts in its scored frames at or after $f_n^{\mathrm{bdry}}$. An action already underway in context cannot be counted again; a later separate onset can count. The response need not land a hit, successfully dodge or intercept, satisfy an exact recovery time, or end at a precise location. Once a valid onset is identified, later extra actions or repetitions do not erase that post-segment success. HNM approach/retreat accepts movement relative to any external player; movement of the other player alone is insufficient.

For a fixed judge $g$ and round $e$, let $b_{n,g,e}^{\mathrm{pre}}\in\{0,1\}$ indicate that every pre-boundary WAIT window passes, and let $b_{n,g,e}^{\mathrm{post}}\in\{0,1\}$ indicate that at least one post-boundary RESPOND window passes. For a split no-op case, $b_{n,g,e}^{\mathrm{post}}=1$ instead requires every post-boundary WAIT window to pass. The case passes only when both indicators are one, giving
\begin{equation}
    \mathrm{JR}_{g,e}=\frac{100}{K_{\mathrm{JR}}}\sum_{n=1}^{K_{\mathrm{JR}}}
    b_{n,g,e}^{\mathrm{pre}}b_{n,g,e}^{\mathrm{post}},
    \qquad K_{\mathrm{JR}}=30.
    \label{eq:eval-joint-reactivity}
\end{equation}
This corresponds to $S_{g,e}^{\mathrm{JR}}$ in
Equation~\ref{eq:eval-case-score}. Cases judged as
\texttt{unknown} receive zero credit and remain in the denominator.

\subsection{Visual Quality (VQ)}\label{apen:eval_vq}

\paragraph{Objective, data, and evidence.}
VQ evaluates visual readability and consistency independently of behavioral correctness on 30 reactivity-evaluation case conditions (from JR) per game. The judge receives the initial identity/appearance reference and three continuous future windows: 1--34, 35--67, and 68--100, with frames 0, 34, and 67 as their respective preceding context frames. Thus, all 100 future frames are covered with 34/33/33 scored frames, and each request has at most 36 images. No action instruction, trigger label, JR score, or future GT reference is provided.

Each window receives four independent integer scores from 0 to 4: clarity ($\mathrm{clar}$), covering readable silhouettes and detail; identity/body structure ($\mathrm{id}$), covering character identity, appearance, and coherent anatomy; background integrity ($\mathrm{bg}$), covering scene structure and layout; and temporal stability ($\mathrm{temp}$), covering continuity across adjacent frames. Clarity and identity/body structure require inspection of every character so that a damaged actor is not hidden by a clean background. Static waiting, native sprite flips, exaggerated poses, legitimate effects, and ordinary game animation are allowed. Incorrect but clearly rendered behavior affects behavior metrics, not VQ. Temporal deductions require evidence from consecutive frames.

\paragraph{Game-specific rubrics.}
HNM assigns 4 for no discernible defect, 3 for a minor/local/brief defect, 2 for an obvious or recurring defect, 1 for severe or persistent degradation, and 0 for an essentially unusable dimension. The released SF3 scores use a more lenient gameplay-demo rubric: 4 permits cosmetic blur, small texture defects, mild ghosting, or minor jitter when fighters and the scene remain readable and continuously trackable. A score of 3 denotes noticeable but generally interpretable artifacts; 2 denotes substantial disruption; 1 denotes severe loss of readability; and 0 denotes unusable evidence. Major identity replacement, body fragmentation, disappearance, or scene destruction still lowers SF3 scores. Therefore, the HNM and SF3 VQ values do not share an identical severity scale, although GT and generated videos within a game use the same rubric.

\paragraph{Aggregation.}
For a fixed judge $g$ and round $e$, we suppress these two indices in the per-video scores. Let $z_{n,\delta,w}^{\mathrm{VQ}}\in\{0,1,2,3,4\}$ denote the score for video $n$, quality dimension $\delta\in\mathcal{D}_{\mathrm{VQ}}=\{\mathrm{clar},\mathrm{id},\mathrm{bg},\mathrm{temp}\}$, and window $w\in\{1,2,3\}$. The scored frame counts are $(\nu_1,\nu_2,\nu_3)=(34,33,33)$, so $\sum_{w=1}^{3}\nu_w=F$. We first weight windows by their scored frame counts, then equally weight the four dimensions and the $K_{\mathrm{VQ}}=30$ videos:
\begin{equation}
    \begin{aligned}
        \overline z_{n,\delta}^{\mathrm{VQ}}
        &=\sum_{w=1}^{3}\frac{\nu_w}{F}z_{n,\delta,w}^{\mathrm{VQ}},\\
        \mathrm{VQ}_n
        &=\frac{100}{16}\sum_{\delta\in\mathcal{D}_{\mathrm{VQ}}}\overline z_{n,\delta}^{\mathrm{VQ}},
        \qquad
        \mathrm{VQ}_{g,e}=\frac{1}{K_{\mathrm{VQ}}}\sum_{n=1}^{K_{\mathrm{VQ}}}\mathrm{VQ}_n.
    \end{aligned}
    \label{eq:eval-visual-quality}
\end{equation}
The denominator $16=4\times4$ accounts for four dimensions, each with a maximum score of four. The resulting 0--100 score is a rubric aggregate, not a success probability. Each judge's three round-level VQ scores are summarized using the same mean/sample-standard-deviation procedure as the behavioral metrics.

\section{Multiple NPCs}\label{apen:multiple_npcs}
\begin{figure}[H]
    \centering
    \includegraphics[width=\linewidth]{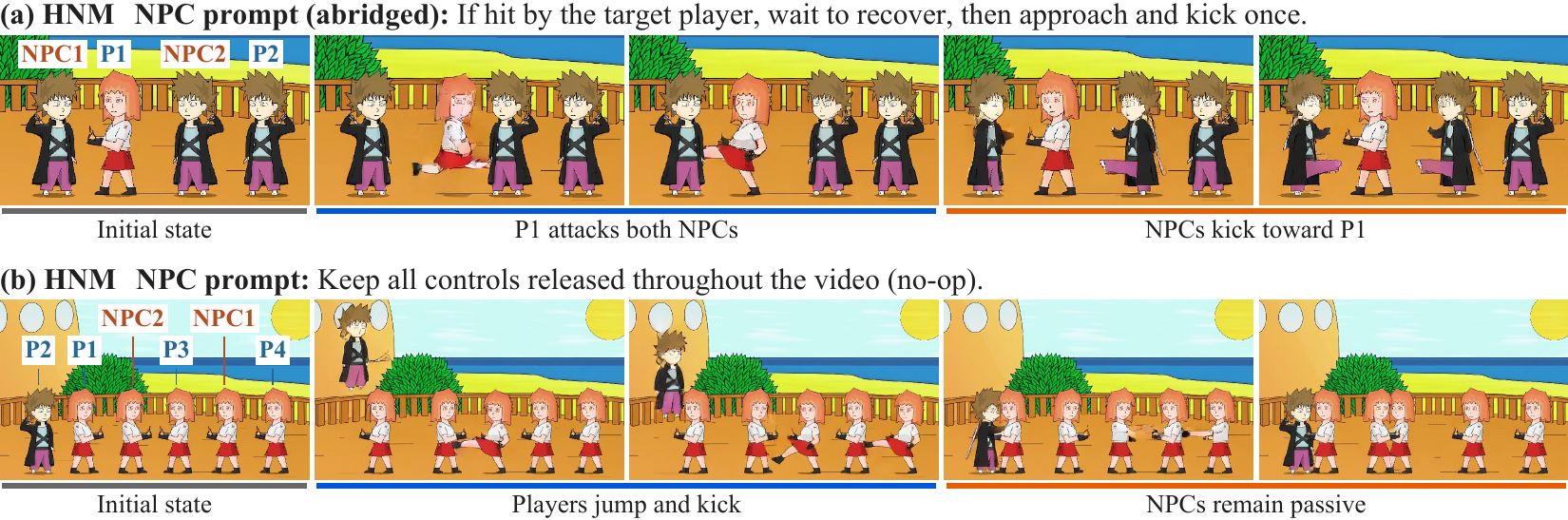}
    \caption{Multiple NPCs generated by \method in HNM. (a) P1 attacks NPC2 and NPC1 in turn; the NPCs then kick toward P1. (b) Two no-op NPCs remain passive while four controlled players act. Each row shows an excerpt from one generated sequence; the prompt applies to both NPCs and is abbreviated in (a). Labels indicate initial roles.}
    \label{fig:multi-npc-examples}
\end{figure}
Figure~\ref{fig:multi-npc-examples} presents two HNM examples with two NPCs in the same generated scene. In (a), a player attacks the NPCs on either side, and both NPCs subsequently kick toward that player. In (b), four players perform different actions while two NPCs retain their passive behavior under no-op instructions. The two NPCs receive the same instruction within each example; their character bindings distinguish which instances the instruction applies to.

\section{Qualitative Results Across Reactivity Types}\label{apen:all_reactivity} 

\begin{figure}[H]
    \centering
    \includegraphics[width=\linewidth]{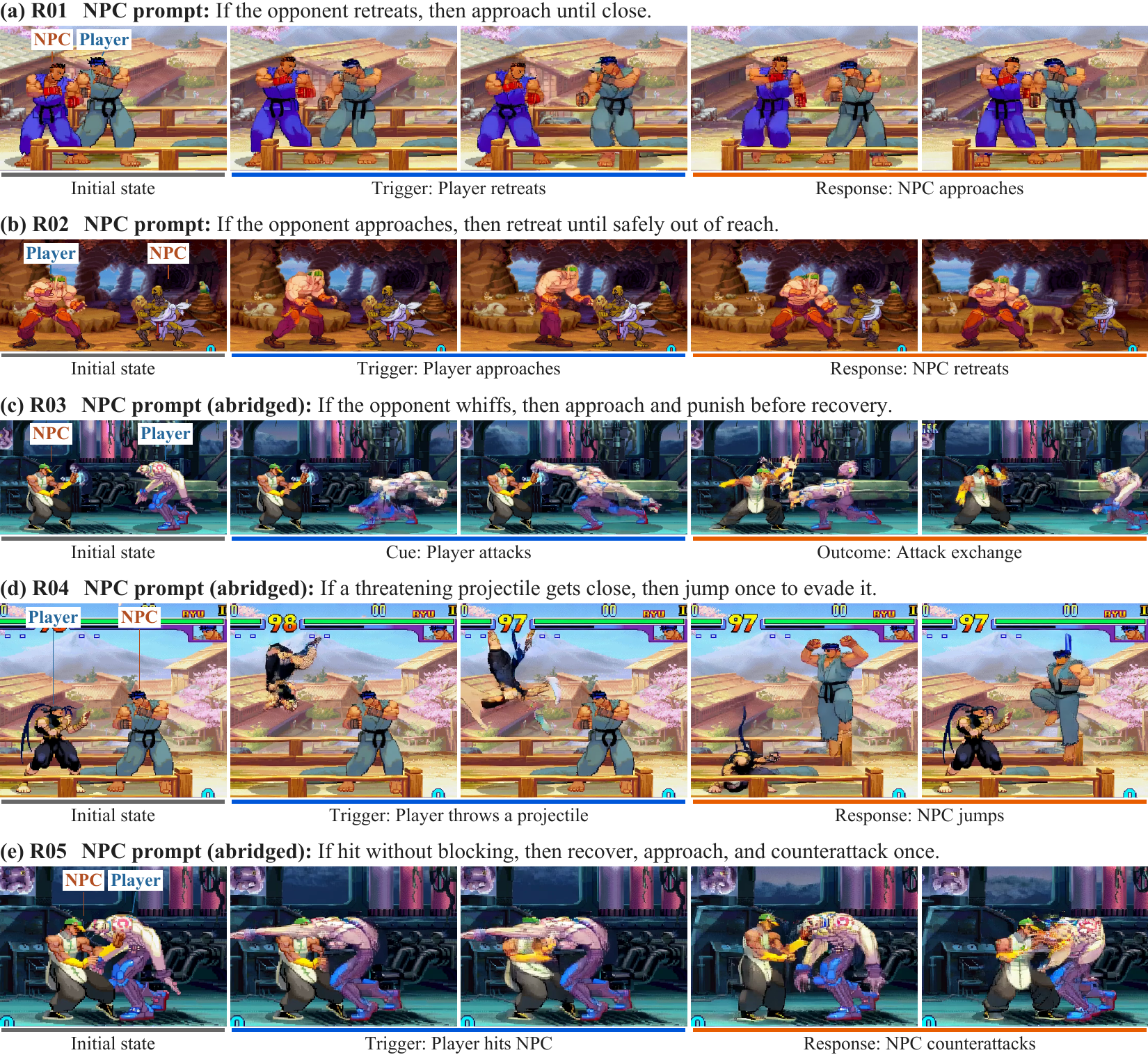}
    \caption{SF3 reactivity examples, R01--R05: approach, retreat, missed-attack punishment, projectile evasion, and counterattack after being hit. The opening event and subsequent behavior are grouped by the blue and orange rules.}
    \label{fig:sf3-all-reactivity-a}
\end{figure}

\begin{figure}[H]
    \centering
    \includegraphics[width=\linewidth]{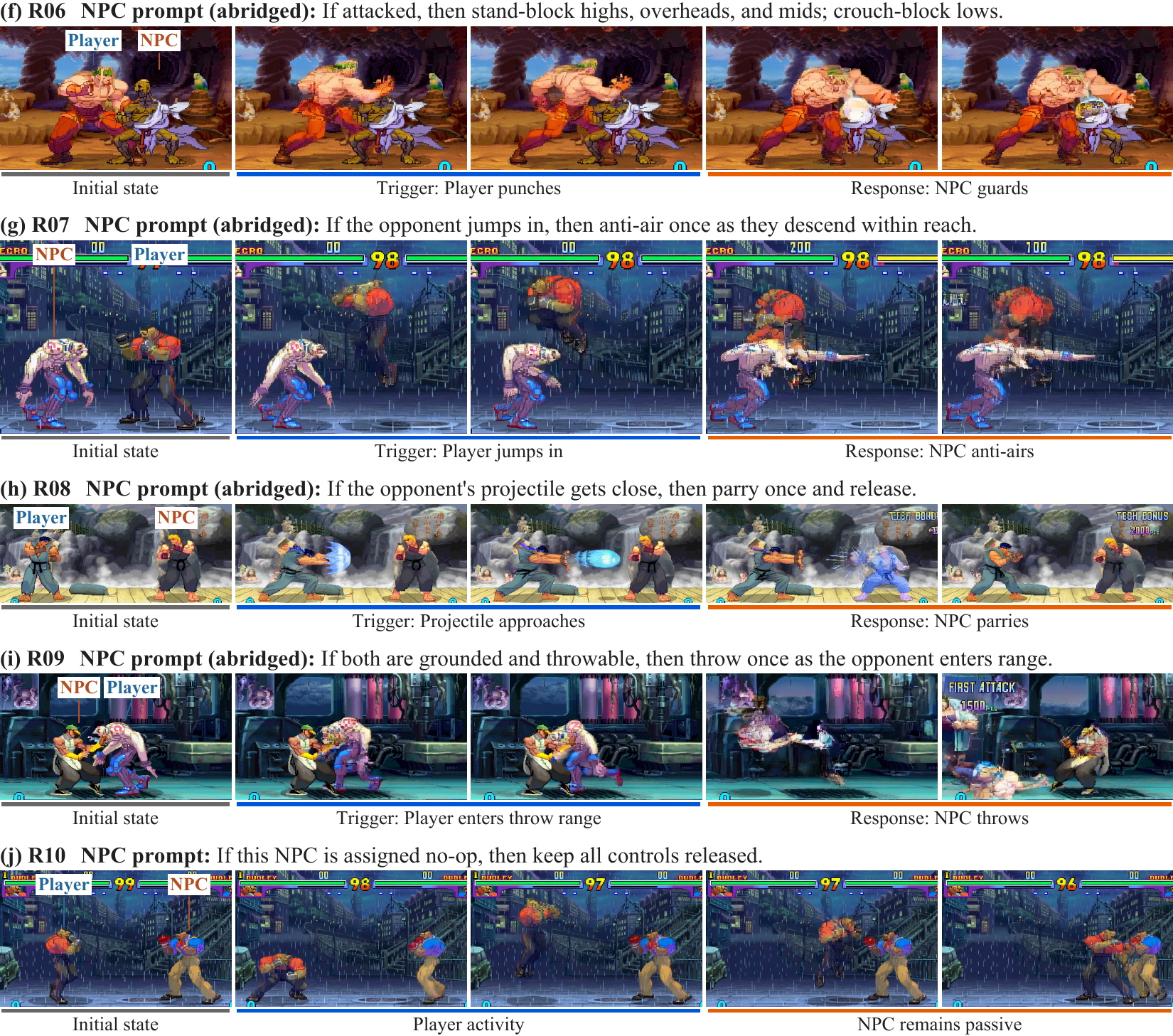}
    \caption{SF3 reactivity examples, R06--R10: guard, anti-air, parry, throw, and no-op. R06 shows a standing guard, and R07 shows an attack against the airborne player. R08 reuses the parry sequence from Figure~\ref{fig:npc-reactivity}(a), showing a blue parry effect followed by a return to stance. R09 shows a throw, while R10 shows the passive NPC during the player's movement.}
    \label{fig:sf3-all-reactivity-b}
\end{figure}

\begin{figure}[H]
    \centering
    \includegraphics[width=\linewidth]{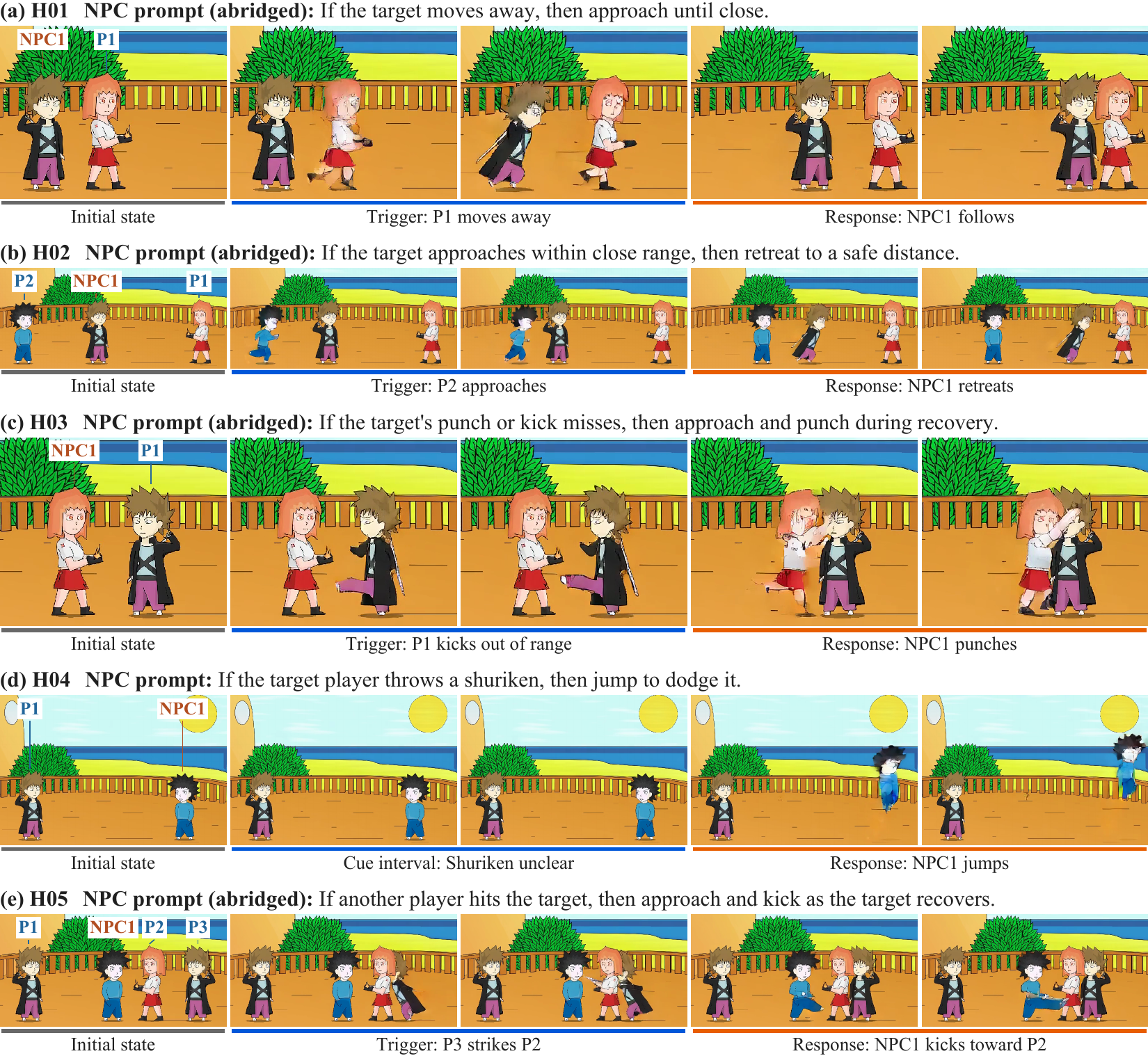}
    \caption{HNM reactivity examples, H01--H05: approach, retreat, missed-attack punishment, shuriken evasion, and response to another player hitting the target. H03 shows an out-of-range kick followed by an NPC punch, without establishing exact recovery timing. In H04 the jump is visible, but the shuriken is not clearly resolved. H05 retains all four characters to show the player interaction preceding the NPC's kick.}
    \label{fig:hnm-all-reactivity-a}
\end{figure}

\begin{figure}[H]
    \centering
    \includegraphics[width=\linewidth]{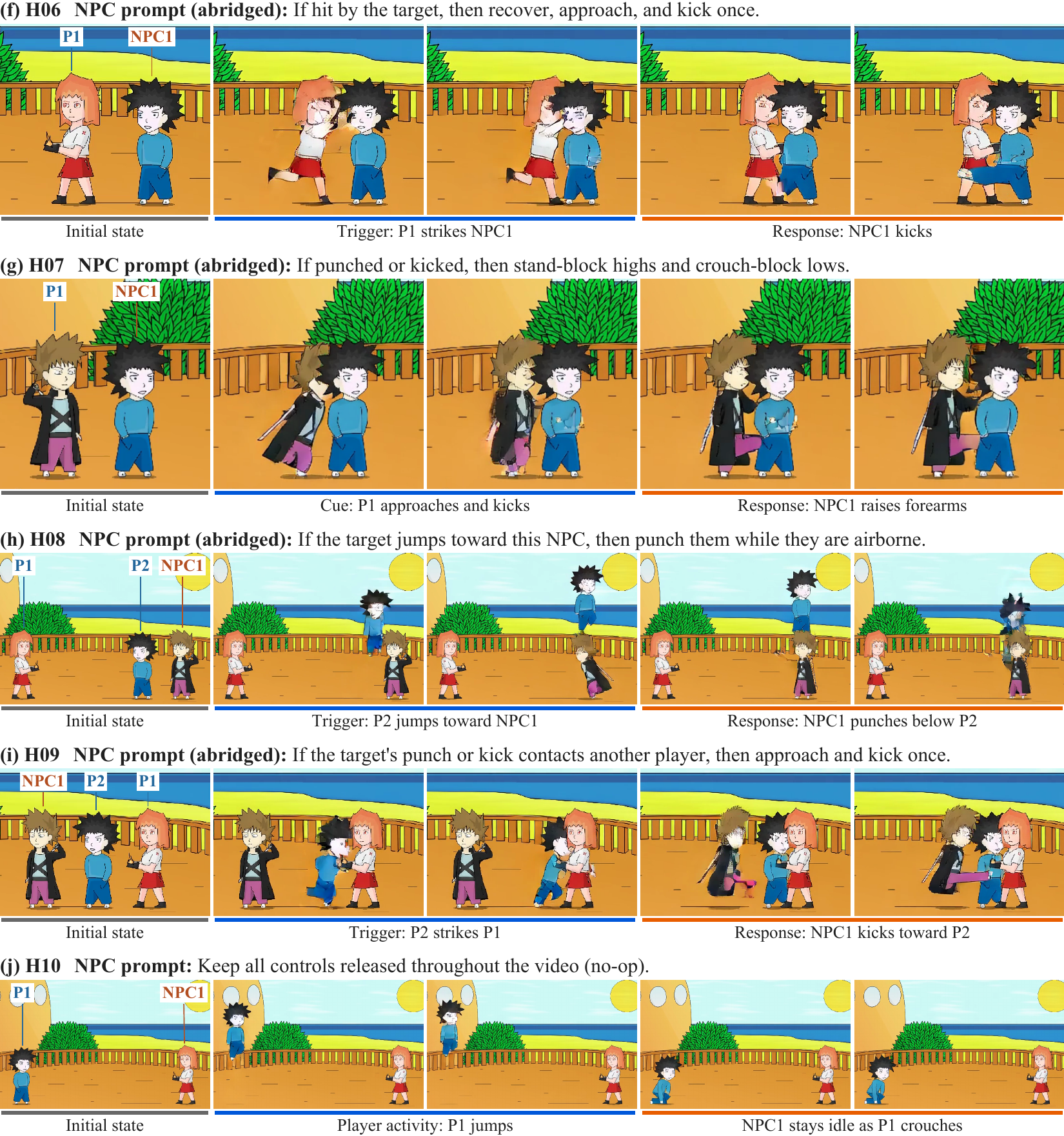}
    \caption{HNM reactivity examples, H06--H10: counterattack after being hit, guard by attack height, anti-air, response to the target hitting another player, and no-op. H07 shows a subtle forearm raise; correct attack-height matching is inconclusive. H08 shows a punch while P2 is airborne, with no clear contact. H09 shows a kick toward P2 after the player interaction, and H10 shows the NPC remaining idle while P1 jumps and crouches.}
    \label{fig:hnm-all-reactivity-b}
\end{figure}

Figures~\ref{fig:sf3-all-reactivity-a}--\ref{fig:hnm-all-reactivity-b} show one generated example for every rule family and the no-op condition in each game. Rule IDs follow Tables~\ref{tab:sf3-rule-catalog} and~\ref{tab:hnm-rule-catalog}. Each row contains five chronological frames: the initial roles, a player event or cue interval, and the subsequent NPC behavior. Prompts marked ``abridged'' summarize the conditioning instruction for better displaying.

\clearpage

\clearpage

\clearpage

\end{document}